\documentclass{article}
\usepackage{spconf,amsmath,graphicx}

\title{Semantic-WER: A Unified Metric for the Evaluation of ASR Transcript for End Usability}
\name{Somnath Roy}
\address{Freshworks Inc. \\
        somnath.roy@freshworks.com}
\begin{document}
%
\maketitle
\begin{abstract}
Recent advances in supervised, semi-supervised and self-supervised deep learning algorithms have shown significant improvement in the performance of automatic speech recognition (ASR) systems. The state-of-the-art systems have achieved a word error rate (WER) less than 5\%. However, in the past, researchers have argued the non-suitability of the WER metric for the evaluation of ASR systems for downstream tasks such as spoken language understanding (SLU) and information retrieval. The reason is that the WER works at the surface level and does not include any syntactic and semantic knowledge. The current work proposes Semantic-WER, a metric to evaluate the ASR transcripts for downstream applications in general. The Semantic-WER can be easily customized for any downstream task.
\end{abstract}
\section{Introduction}
\label{sec:intro}

Speech recognition systems back in early 2000 were mainly HMM-based models \cite{young2002htk}. In the last two decades, the entire landscape has changed for machine learning and deep learning in general and speech recognition in particular. We now have access to thousands of hours of annotated speech databases in multiple languages \cite{ardila2019common}. There are many open-source toolkits available for developing an ASR system \cite{povey2011kaldi, ravanelli2019pytorch, watanabe2018espnet, wang2019espresso, shao2020pychain, baevski2020wav2vec}. The popular architectures for getting good performance are DNN-HMM \cite{vesely2013sequence}, LSTM-RNN \cite{graves2013hybrid}, time-delay neural network \cite{povey2016purely} and CNN \cite{palaz2015analysis}. Nowadays, most of the end-to-end ASR systems use the popular architecture called transformer \cite{vaswani2017attention, baevski2020wav2vec}. 
Despite such a giant leap, the means of evaluating the quality of a speech recognition system has remained mostly unchanged. WER is still the de facto standard metric for ASR system assessment. It is calculated by the total error count normalized by the reference length (N$_r$), as shown in equation 1. The total error count (i.e., the sum of substitutions(S), insertions(I), and deletions(D)) is computed by performing the Levenstein alignment for reference and hypothesis word sequences. \\

\begin{equation}
    WER = \frac{S+D+I}{N_r}
\end{equation}
WER is a direct and objective measure for evaluating the quality of ASR transcripts. However, there are certain limitations of WER and its application to end usability \cite{he2011word, favre2013automatic, wang2003word, mccowan2004use, morris2004and}. The main two limitations are stated below. 
\begin{itemize}
    \item The numerator in WER is not bounded by the length of reference because of the inclusion of insertions in the total number of edits. Therefore, the normalization by N$_r$ is not bounded to [0, 1].
    \item Both content words and function words are equally important, not ideal for most downstream tasks. 
\end{itemize}
The applicability of WER for downstream uses may not be straightforward. The following examples further demonstrate the fuzziness of WER for its relevance to the downstream tasks.

\begin{itemize}
    \item Ref: f*ck you
    \item Hyp1: thank you
    \item Hyp2: okay
\end{itemize}
Hyp1 and Hyp2 are hypotheses from two ASR models in the above example, and Ref denotes the reference transcript. The meaning of Hyp1 and Hyp2 utterly opposite to the meaning of the Ref. However, the WER fails to capture these semantic differences and assigns a score of 0.5 and 1.0 to Hyp1 and Hyp2, respectively. It implies that low WER may not be a good indicator for better sentiment analysis accuracy.  

\begin{itemize}
    \item Ref: My name is harvey spelled as h. a. r. v. e. y. 
    \item Hyp1: My name is hurdy spelled as empty
    \item Hyp2: My name is hurdy spelled as age a. r. v. e. y.
\end{itemize}
The 'empty' word in Hyp1 denotes the blank or no output. Both the hypotheses cannot capture the name accurately. However, the Hyp2 is better in capturing the spelled out for the proper noun. Considering each spelled letter as a word results in a WER of 0.58 and 0.16 for Hyp1 and Hyp2, respectively. However, one can argue that the spelled letters should combine to form a single word before WER. In that case, the WER for Hyp1 and Hyp2 would be 0.28 and 0.37.  Such inconsistent scoring is far from reliable for spoken language understanding \cite{wang2003word}.  \\

The motivation for the present work is stated below.
\begin{itemize}
    \item SlotWER is computed alongside the WER for the evaluation of SLU systems \cite{liu2021domain}. However, we need a unified evaluation framework that is direct and objective and can be used for accessing the quality of the ASR transcript for end usability.
    \item In the conversational domain, a single word or phrase may determine the overall sentiment of a transcript; therefore, such words should have higher weight while scoring. The present work decides weight by the semantic distance and the effect of a miss of an entity on the entire utterance.
    \item The weight can also be configured for particular words or phrases where semantic distance is not effective in general.
    \item Many organizations using ASR transcripts have access to customer relationship management (CRM) data for lookup or post-processing. In such cases, we should allow a certain level of discounting for spelled-out entities. In other words, if the ASR transcript misses one or two-character, e.g., "h a r v e y" is transcribed as "h r v e y", then it should not be penalized much.
    \item Finally, the metric must be bounded in the range of [0, 1].

\end{itemize}
In this work, we propose an alternative evaluation metric called Semantic-WER, which leverages the benefits of WER and augments it with the semantic weight of a word according to its importance to the end usability as described above. The user can easily customize the proposed system according to the downstream tasks at hand. \\

The remainder of this paper is organized as follows. Section 2 describes the related works. Section 3 describes the Semantic-WER, and its effectiveness for the downstream tasks in general. In Section 4 we validate the proposed metric by finding it's correlation with the human errors. The conclusion and limitation are described in Section 5.

\section{Related Works}
\label{sec:RW}

A weighted WER is proposed in \cite{hunt1990figures, hunt1988evaluating} to avoid the bias in calculating the total error due to dynamic programming alignment. The bias is avoided by reducing the weight of insertion and deletion by a factor of 2. A similar approach is followed in \cite{makhoul1999performance, mccowan2004use} for avoiding the alignment bias. 
An alternative metric called Word information lost (WIL) proposed in \cite{wang2003word} effectively bounds the error in the range of [0, 1] as shown in equation (2) and equation (3). However, the normalization by both the reference length (N$_r$) and hypothesis length (N$_h$) as shown in equation (2) can only be effective for the cases having hypothesis length longer than the length of a reference. Moreover, WIL does not use any syntactic or semantic knowledge-based weight to penalize the miss of essential words. On a similar information theoretic standpoint  \cite{mccowan2004use} proposes that precision and recall can be a better alternative to WER. However, precision and recall also have the same limitations and do not include syntactic or semantic knowledge. A work that closes the proposed work is \cite{garofolo19991998}.  It presents three additional metrics for information retrieval end usability. These metrics are named entity word error rate (ne-wer), general IR-based, and query-word word-error-rate(qw-wer). The general IR based metric has three components- i) stop-word-filtered word error rate (swf-wer), ii) stemmed stop-word-filtered word error rate (sswf-wer), and iii) IR-weighted stemmed stop-word-filtered word error rate (IRW-WER).
\begin{equation}
    WIP = \frac{H}{N_r}\frac{H}{N_h} = \frac{I(X, Y)}{H(Y)}
\end{equation}

\begin{equation}
    WIL = 1 - WIP
\end{equation}

\section{Semantic-WER}
\label{sec:sw}

Semantic-WER (SWER) is a direct and objective measure similar to WER. Unlike WER, SWER has specific weights for substitution, deletion, and insertion. The substitution weight (W$_{sub}$), as shown below in equation (4), has four cases.

\small
\begin{equation}
  W_{sub} =
    \begin{cases}
      1, \text{if r$_w$} \in \text{${NE}$} \cup  \text{${SENT}$}\\
      cer(r_{w}, h_{w}), \text{if r$_w$} \in \text{${SE}$}  \\
      1, \text{if similarity(r$_w$, h$_w$) $<$ 0.6  and r$_w$}  \notin \text{${NE}$} \cup  \text{${SENT}$}  \\
      0, \text{if similarity(r$_w$, h$_w$) $>$ 0.6 and r$_w$} \notin \text{${NE}$} \cup  \text{${SENT}$}\\
    \end{cases}
\end{equation}

 ${NE}$ $\cup$ ${SENT}$ represents a set having named entities and sentiment words. r$_w$ and h$_w$ denote the reference word and hypothesis word, respectively. The first case does not differentiate between short and long utterances having words belonging to ${NE}$ $\cup$ ${SENT}$. In other words, both long and short utterances are penalized equally for error. $SE$ represents the spelled out entities. It is ubiquitous in telephonic conversation to spell out the entities to the listener. Therefore, the second case uses character error rate (cer) for spelled out entities. Most of the organizations which analyze telephonic conversations have access to CRM data. Therefore, one can have a threshold to relax the number of characters' substitution for the spelled-out entities. Current work sets the number of character substitution thresholds to zero, i.e., even a single character substitution counts. Unlike \cite{garofolo19991998}, the last two instances use the cosine similarity score between the embedding of r$_w$ and h$_w$ for obtaining the substitution penalty. 
Such similarity-based scores have the following unique advantages. 
\begin{itemize}
    \item It yields a high score for semantically similar words and therefore assigns zero substitution penalty. e.g., r$_w$ = "go" and h$_w$ = "goes".
    \item It yields low score semantically dissimilar words and therefore assigned a substitution penalty of one, e.g., r$_w$ = "tortoise" and h$_w$ = " rise".
    \item The above two benefits allow us to use all words without any pre-processing (i.e., lemmatization and stop word removal) as proposed in \cite{garofolo19991998}.
\end{itemize}
Similar to W$_{sub}$, the deletion weight (W$_{del}$) has three cases, as shown in equation (5). The first two cases are the same as of W$_{sub}$.
\small
 \begin{equation}
     W_{del} = 
     \begin{cases}
        1,  \text{if r$_w$} \in \text{${NE}$} \cup  \text{${SENT}$} \\
         cer(r_{w}, h_{w}), \text{if r$_w$} \in \text{${SE}$}  \\
        \frac{1}{N_{r}}, \text{otherwise} 
     \end{cases}
 \end{equation}
 
Unlike \cite{mccowan2004use, hunt1990figures}, the third case assigns a weight of $\frac{1}{N_{r}}$ to deletion for rest of the words. Finally, the insertion weight W$_{ins}$ is equal to the probability mass distributed over the hypothesis, i.e., $\frac{1}{N_h}$ as shown in equation (6).
\small
\begin{equation}
  W_{ins} = \frac{1}{N_{h}} 
\end{equation}
The normalization by hypothesis length (N$_h$) rather than N$_r$ is because the insertions take place in the hypothesis. Following are the reasons for assigning such a small weight to the insertion penalty.
\begin{itemize}
    \item The objective of an ASR system is to transcribe the input speech. However, in most practical cases, it has been observed that the insertions are due to not having a better voice activity detection (VAD) or speech activity detection (SAD) system. Therefore, it is better not to penalize the ASR system for the VAD errors.
    \item Some overlapped or non-comprehensible speech part which is not transcribed in the reference; however, the ASR system transcribes it.
    \item It is also likely that the human transcribers have missed some words in the reference which ASR systems are correctly capturing.
    \item Given that the acoustic models are trained on the thousands of hours of data, the insertions may be due to not fusing the in-domain text in the language model.  
\end{itemize}

An intermediate score called score$_{a}$ aggregates the substitution, deletion, and insertion errors as shown below in equation (7). 
\small
\begin{equation}
    score_a = \sum\limits_{S, D \notin NE\cup SENT} W_{sub}*S + W_{del}*D + W_{ins} * I \\
\end{equation}

\subsection{Importance Weight and Distributed Weight}
Entities and sentiment words can be pretty important as well as sensitive in a transcription. The importance weight (IW) is based on the assumption that a wrong entity can maximally affect all other entities in a sentence. In contrast, a wrong sentiment word can change the semantics of the entire sentence or a phrase. Therefore, the IW lies between 1 to the maximum number of entities present in an utterance for a wrong entity. And the IW of a wrong sentiment word lies between 1 to the maximum number of words in a phrase. The following equations (9) and (10) compute the distributed weight (DW) and the SWER respectively. \#E$_{(NE \cup SENT)}$ denotes the number of wrong entities and sentiment words in an utterance.
\small
\begin{equation}
  accuracy = 1 - score_a
\end{equation}
\small
\begin{equation}
DW = \frac{accuracy}{N_r - \#E_{(NE \cup SENT)}}
\end{equation}

\small
\begin{equation}
  SWER = score_a + DW \times IW
\end{equation}

Three example sentences having the same WER and different SWER demonstrate the effectiveness of SWER, as shown below in Table 1. 
\begin{table}[h]
\addtolength{\tabcolsep}{-0.0pt}
\begin{center}
\begin{tabular}{ c c c }
 \hline
 {Ref and Hyp} & {WER} & {SWER}\\
 \hline
 {what did \textbf{you} do in \textbf{paris}} \\
 {what did \textbf{u} do in \textbf{phariz}} & {0.33}  & {0.46} \\
 \hline
 {i love \textbf{switzerland}} \\ {i love \textbf{switjerlan}} & {0.33} & {0.66} \\
 \hline
 {ram \textbf{loves} sita} \\ {ram \textbf{love} sita} & {0.33} & {0.0}\\
 \hline 
\end{tabular}
\end{center}
\caption{Example sentences for comparison of WER and SWER. The value of SWER is computed using the default importance weight i.e., IW = 1.}
\end{table}

\section{Validation}
\label{sec:typestyle}
The CoNLL-2003 NER dataset marks the begining of a new topic by "-DOCSTRART". It is used for extracting sentences from each domain to capture a better distribution in terms of unique named entities. Only two sentences are extracted from each domain. A total of 946 sentences extracted for the validation purpose. These sentences are labelled for categories like person (PER), organization (ORG), location (LOC) and miscellaneous (MISC) at word level. A sample sentence with it's label is shown below in Table 2. 
\begin{table}[h]
\addtolength{\tabcolsep}{-3pt}
\begin{center}
\begin{tabular}{ |c|c|c|c|c|c|c|c|c| }
 \hline
{EU} & {rejects} & {German} & {call} & {to} & {boycott} & {British} & {lamb} & {.} \\\hline
{I-ORG}&  {O}  & {I-MISC} & {O} & {O} &  {O} & {I-MISC} &  {O}  &  {O}\\\hline
\hline
\end{tabular}
\caption{Sample CoNLL-2003 sentence labeled for all named entities. }
\end{center}
\end{table}
The detailed description of the dataset-NER can be found in the Table 3. All 946 sentences are synthesized at a sampling rate of 16KHz using an end-to-end Text-to-Speech Synthesizer \cite{watanabe2018espnet}. 
\begin{table}[h]
\addtolength{\tabcolsep}{-2pt}
\begin{center}
\begin{tabular}{ |c|c| }
 \hline
{\# Unique Words} &{2500} \\ 
\hline
{\# I-ORG} & {529} \\
\hline
{\# I-LOC} & {221} \\
\hline
{\# I-PER} & {456} \\ 
\hline
{\# I-MISC} & {637} \\ 
\hline
\end{tabular}
\caption{The distribution of category wise named entities in the dataset-NER }
\end{center}
\end{table}
\\
 
 \subsection{SUBJECTIVE SCORING}
The sentences from the dataset-NER are categorized into three categories, namely Cat-I, Cat-II, and Cat-III, based on the number of entities. Cat-I, Cat-II, and Cat-III consist of sentences having one, two, and three entities respectively. Cat-I has 354, Cat-II consists of 344, and Cat-III contains 146 sentences. A total of sixty sentences are selected by randomly extracting twenty sentences from each category. A perceptual experiment was designed using the Praat MFC experiment. Each audio stimuli has several text responses. The number of responses equals two times the number of entities (i.e., one for substitution and one for deletion) present in the reference text as shown above in Fig. 1. Since it is pretty unusual to consider the insertion of an entity or sentiment word, we deliberately kept the insertion error out of the experiment. A total of 10 participants rated each response on a scale of 1 (poor) to 5 (excellent). The subjective accuracy of each response is calculated by averaging the responses for each stimulus and then dividing it by the highest score i.e., 5. Finally, the human assigned WER (HWER) is calculated as one minus the subjective accuracy. 
 
 \begin{figure}
\includegraphics[width=1.0\linewidth]{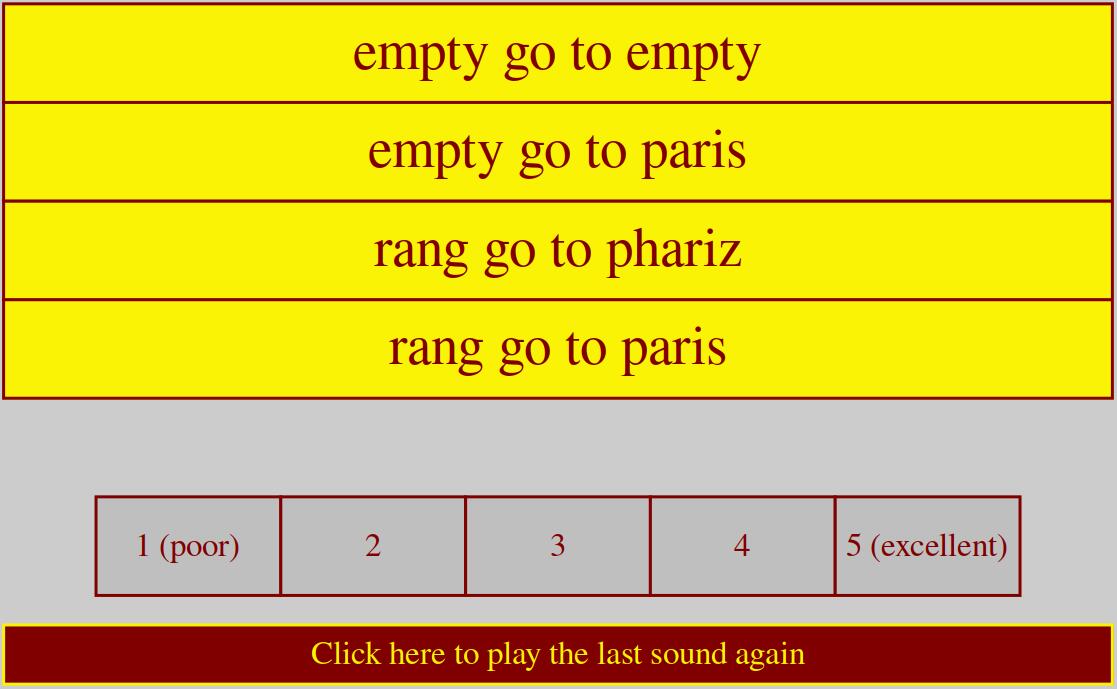}
\caption{Praat MFC Experiment for Subjective Scoring for reference "ram goes to paris". The top two response stumli are hypotheses post deletion and bottom two are for substitution error. "empty" word is used to show the deletion of a word.}
\end{figure}
 
 \begin{figure}
\includegraphics[width=1.0\linewidth]{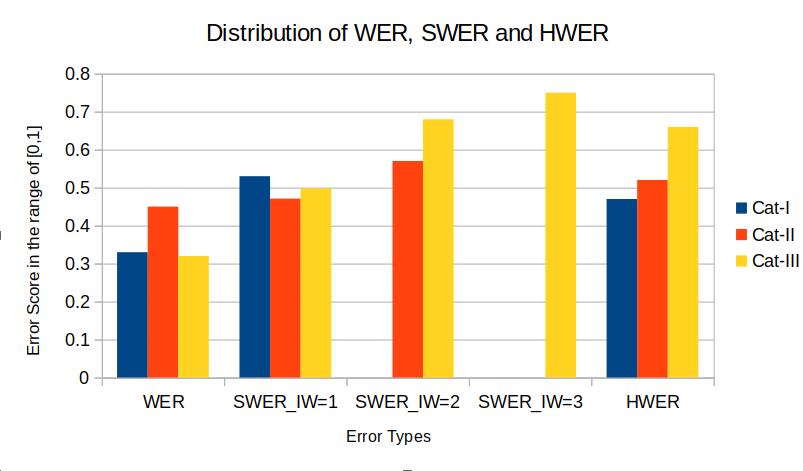}
\caption{The distribution of errors (WER, SWER and HWER) per category. Cat-I, Cat-II and Cat-III denote the sentences in which one, two and three named entities are substituted/deleted respectively.}
\end{figure}

\subsection{Correlation of WER, SWER and HWER}
\label{sec:majhead}
 The difference between WER and HWER is significantly higher than between SWER and HWER, as shown in Fig 2. The SWER\_IW=1, SWER\_IW=2, SWER\_IW=3 represent the SWER with importance weight of 1, 2, and 3, respectively. The correlation of WER and HWER is less than 0.7 across categories. However, the correlation of WER and SWER is mainly in the range of 0.7 to .85. The SWER\_IW=1 has the highest correlation with HWER followed by SWER\_IW2. It implies that the default importance weight is good for sentences with one entity and the importance weight of two for the rest of sentences. The distribution of correlation of errors with HWER across categories is shown in Fig 3. 
 
 \begin{figure}[h]
\includegraphics[width=1.0\linewidth]{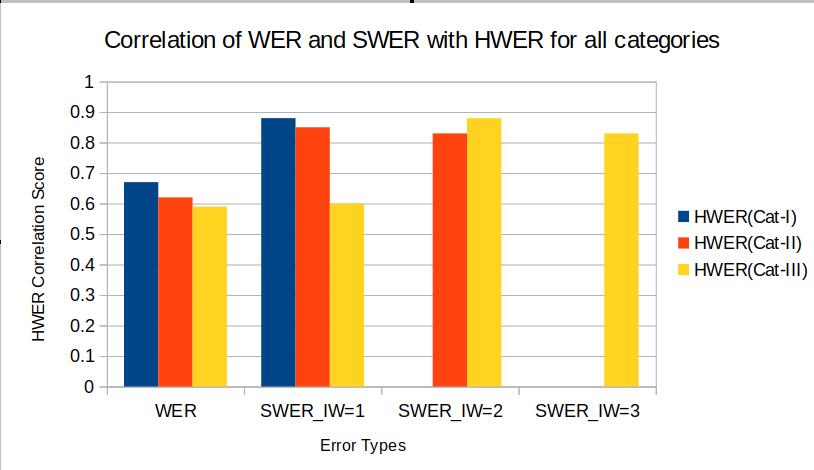}
\caption{Correlation of WER, SWER\_IW=1, SWER\_IW=2 and SWER\_IW=3 with HWER for all three categories.}
\end{figure}

\vfill\pagebreak

\section{CONCLUSION and LIMITATIONS}
Entities, in general, are challenging to recognize compared to non-entity words. Also, some entities are harder to recognize compared to other entities. One can find a helpful benchmark describing the accuracy of different ASR systems for entities in \cite{del2021earnings}. Librispeech and Switchboard corpora are old corpora and not good enough to further track the progress in the field of speech recognition \cite{chen2021gigaspeech}. On a similar note, we advocate the urgent need for a new metric capable of penalizing more for wrong entities than other common non-entity words. The SWER metric is a proposal for the same and correlates better with the subjective score (HWER) than WER. One limitation of SWER i.e., the reference must be tagged for the entities to be evaluated.   

\bibliographystyle{IEEEbib}
\bibliography{strings,refs}

\end{document}